\documentclass{article}
\usepackage{spconf,amsmath,graphicx,amsfonts,amssymb,bm}
\usepackage{float}
\usepackage{tabu}                     
\usepackage{multirow}                 
\usepackage{multicol}                 
\usepackage{multirow}                
\usepackage{makecell}                 
\usepackage{booktabs}                 

\usepackage{hyperref}
\hypersetup{
    colorlinks=true,
    linkcolor=blue,
    filecolor=magenta,      
    urlcolor=magenta,
    pdftitle={Overleaf Example},
    pdfpagemode=FullScreen,
    }

\title{A Convolutional-Transformer Network for Crack Segmentation \\ with Boundary Awareness}
\name{Huaqi Tao$^{1, *}$, Bingxi Liu$^{2, 3, *}$\thanks{* Equal contribution.}, Jinqiang Cui$^{2}$, Hong Zhang$^{3, \dag}$\thanks{\dag Corresponding author.}, Fellow, IEEE}
\address{$^{1}$Guangxi University, Nanning, China\\
         $^{2}$Peng Cheng Laboratory, Shenzhen, China\\
         $^{3}$Southern University of Science and Technology, Shenzhen, China}
\begin{document}
	%
	\maketitle
	\begin{abstract}
		Cracks play a crucial role in assessing the safety and durability of manufactured buildings. However, the long and sharp topological features and complex background of cracks make the task of crack segmentation extremely challenging. In this paper, we propose a novel convolutional-transformer network based on encoder-decoder architecture to solve this challenge. Particularly, we designed a Dilated Residual Block (DRB) and a Boundary Awareness Module (BAM). The DRB pays attention to the local detail of cracks and adjusts the feature dimension for other blocks as needed. And the BAM learns the boundary features from the dilated crack label. Furthermore, the DRB is combined with a lightweight transformer that captures global information to serve as an effective encoder. Experimental results show that the proposed network performs better than state-of-the-art algorithms on two typical datasets. Datasets, code, and trained models are available for research at \href{https://github.com/HqiTao/CT-crackseg}{https://github.com/HqiTao/CT-crackseg}.
	\end{abstract}
	\section{Introduction}
	\label{sec:intro}
	
	Crack is a prevalent early defect in manufactured buildings. Even tiny cracks can serve as an early warning sign of a potentially serious accident in certain buildings, such as airport runways, bridge decks, and nuclear power plants. As a result, it is crucial to detect the existence of cracks early and repair them promptly. In practical applications, a pixel-level method is required to obtain the crack location and geometric features (i.e., width, length, and extension path). Therefore, most researchers \cite{zhou2022tunnel,konig2019convolutional,yang2019feature,liu2019deepcrack} consider extracting crack information as a semantic segmentation task in computer vision.
	
	Crack segmentation aims to provide a binary segmentation of the data, namely crack and non-crack. Previous research has shown the high accuracy of convolutional neural networks (CNNs), especially fully convolutional neural networks (FCNs) in crack segmentation \cite{konig2019convolutional,zhang2016road,zou2018deepcrack}. Furthermore, researchers have explored various types of convolutional layers to enhance the performance of CNNs for crack segmentation. For example, \cite{zhou2022tunnel} and \cite{zhou2022lightweight} utilize dilated convolution \cite{yu2015multi} in their methods and get better performance. Inspired by these, we can take advantage of hybrid dilated convolution \cite{wang2018understanding} and deformable convolution \cite{zhu2019deformable} to improve the network performance. However, CNNs usually exhibit limitations in modeling explicit long-range relationships due to their intrinsic locality\cite{dosovitskiy2020image} and most cracks show elongated, narrow structures. Therefore, it is essential to get both local and non-local information for accurate crack segmentation.
	
	Vision transformers (ViTs) \cite{dosovitskiy2020image} have recently emerged as a paradigm of deep learning models, which can extract and integrate global contextual information based on a self-attentive mechanism. Many works have proposed convolutional-transformer (CViT) architectures for image segmentation, such as TransUNet \cite{chen2021transunet}, TMUnet \cite{azad2022contextual}, which can leverage both local information from CNN features and global information encoded by Transformers. Although CViT-based models can perform better than CNN-based models in terms of the accuracy of image segmentation, their inference time is longer due to ViTs being heavier than CNNs. And crack segmentation tasks are typically performed on mobile devices, such as drones and phones, which have limited computing resources, memory space, and energy supply. Therefore, we aim to explore the use of a lightweight ViT, such as MobileViT \cite{mehta2021mobilevit}, to design a network that can improve crack segmentation accuracy with as few parameters as possible.
	 
	\begin{figure*}
		\centering
		\includegraphics[width=1.0\textwidth,scale=1.00]{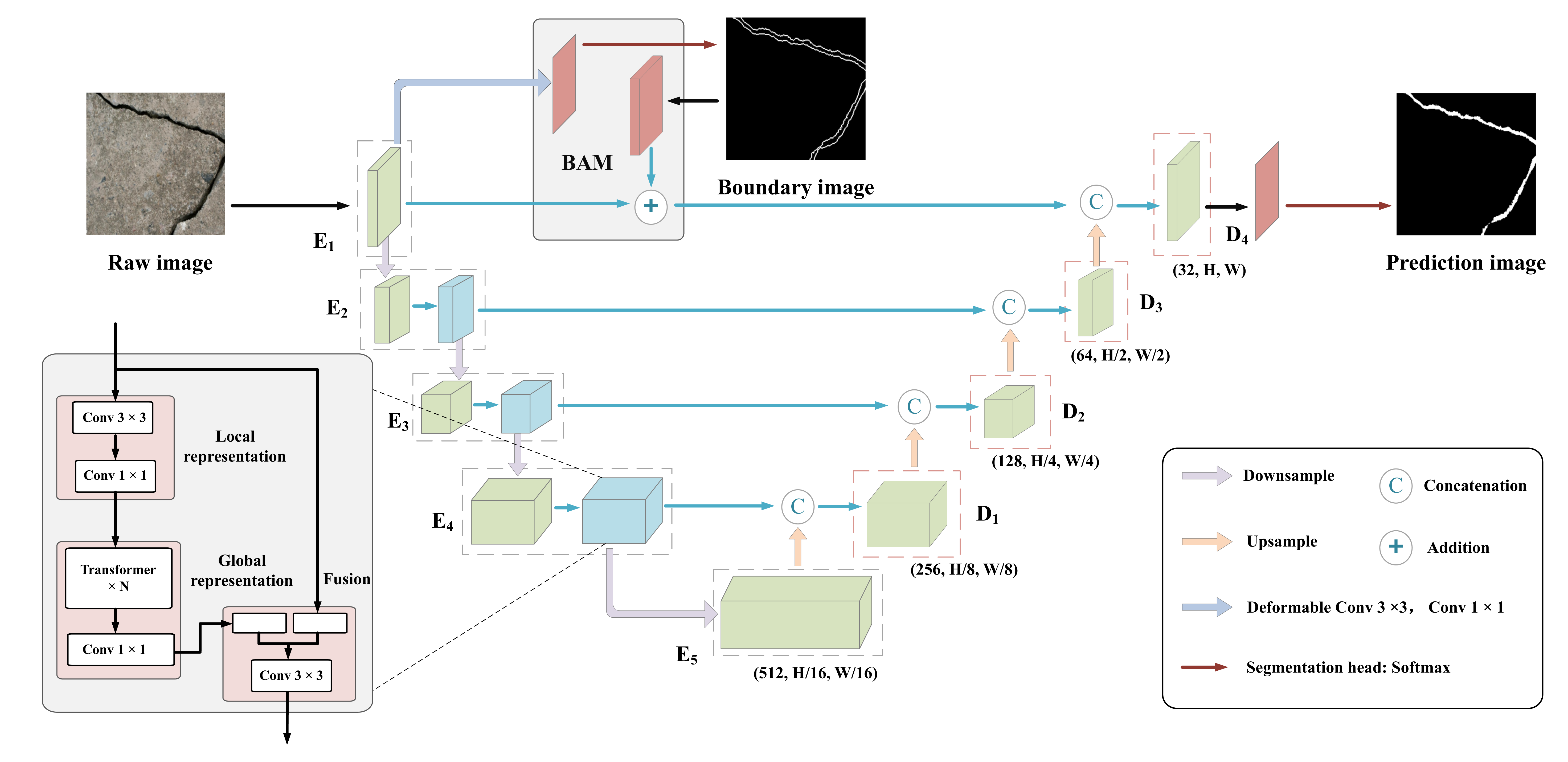}
		\caption{An overview of the proposed model. E and D denote encoder and decoder respectively. Green feature maps are generated via Dilated Residual Block while blue feature maps are generated via MobileViT Block.}
		\label{fig:mymodel}
	\end{figure*}
	
	In this paper, we propose a convolutional-transformer network based on encoder-decoder architecture (U-Net \cite{ronneberger2015u}) to tackle the above issues. The main contributions of our work are summarized as follows: (1) We propose a novel convolutional-transformer network for crack segmentation, which makes full use of local and global information. (2) Two convolutional components are proposed to obtain local information, including a Dilated Residual Block (DRB) and a Boundary Awareness Module (BAM). (3) An effective encoder based on DRB and MobileViT Block is designed to capture global information and keep low parameters. (4) Experimental results show that our method achieves state-of-the-art performance on two benchmark datasets, and the proposed modules can improve the performance in the ablation experiment. Moreover, our source code is shared with the community to promote the research of crack segmentation.

	\section{Method}
	\label{sec:method}
 
	Our proposed network based on encoder-decode architecture is shown in Fig.\ref{fig:mymodel}. It contains three major components: Dilated Residual Block, Boundary Awareness Module, and MobileViT Block. Adjacent encoders are connected by max-pooling operation, which downsamples the feature map by a factor of two. Correspondingly, each decoder is connected by an upsampling operation which doubles the resolution of the feature map. Moreover, the upsampling operation in our network is dense upsampling convolution (DUC) \cite{wang2018understanding} instead of conventional transposed convolution to avoid the detail of crack missing in bilinear upsampling. More details of our network are discussed in the following sections.
	
	\subsection{Dilated Residual Block}
	\label{ssec:DRB}
	
	Dilated convolutions improve the network performance by a larger receptive field, but their receptive field is discontinuous. And it is adverse to extract crack features since most cracks are only two or three pixels wide. Hybrid dilated convolution \cite{wang2018understanding}, in which receptive field can cover a square region without any holes, solve this problem well through a series of dilated convolutions with particular expansion rates. Therefore, we design dilated residual block (DRB) based on a hybrid dilated convolution framework to replace conventional convolutional layers. With the larger receptive field, DRB can learn higher-level semantic features and clearer local crack features.
 
\begin{figure}[t]
		\centering
		\includegraphics[width=\linewidth,scale=1.00]{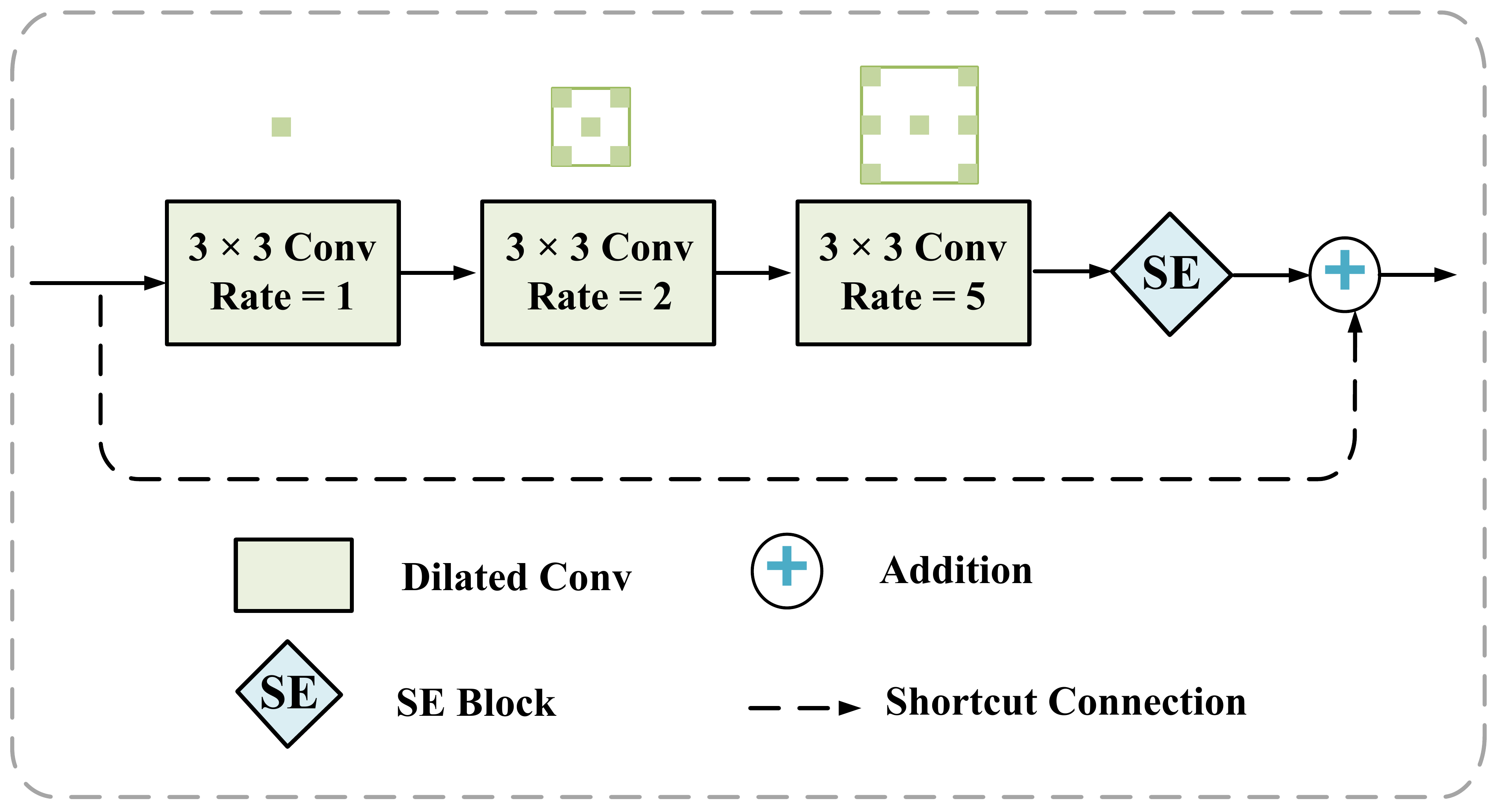}
		\caption{Diagram of the dilated residual block (DRB).}		
		\label{DRB}
	\end{figure}

    As is shown in Fig.\ref{DRB}, DRB consists of three $3 \times 3$ dilated convolutions and a squeeze-excitation (SE) block \cite{hu2018squeeze}. Three dilated convolutions with the expansion rate of 1, 2, and 5 in order to form a hybrid dilated convolution framework to obtain local information. The first dilated convolution can change the feature map channels, whereas the other two do not. The SE block can obtain channel attention. At the end of DRB, the residual connection \cite{he2016deep} is applied to address the vanishing gradient problem and accelerate convergence.

 	\subsection{Boundary Awareness Module}
	\label{ssec:BAM}
 
	Due to the curve feature of cracks and the brightness difference between the boundary and middle of cracks, it is difficult to distinguish the crack boundary from the background under various disturbances. Deformable convolutions have decent performance in learning variable geometric features because of their strong spatial adaptability. Therefore, we employ deformable convolution to design the boundary awareness module (BAM) as shown in the top gray box in Fig. \ref{fig:mymodel}. 
 
    To learn finer boundary features, we create the boundary label from the crack's ground truth by image dilation. The label leads BAM to focus on the crack boundary rather than just the crack. Since more detailed boundary information is retained in the low-dimensional feature stage \cite{liu2019deepcrack}, we put BAM in the shortcut connection between $\rm{E_1}$ and $\rm{D_4}$.
    
    Given a feature map  $x_1 \in \mathbb{R}^{H \times W \times 32}$ from $\rm {E_1}$ with spatial dimension $H$ and $W$, the boundary feature map $B \in \mathbb{R}^{H \times W \times 1}$ can be formulated as follows

     \begin{equation} \label{eq:1}
		  B = \sigma ({\rm Conv_1} ({\rm Dconv}(x_1))),
    \end{equation}
    where $\rm Conv_1$ and $\rm Dconv$ denote $1 \times 1$ convolution and $3 \times 3$ deformable convolution, respectively, and $\sigma$ is Sigmoid activation. Then, we reshape the crack boundary feature map $B$ to $B' \in \mathbb{R}^{H \times W \times 32}$. Thus, the final output $x_1'$ of the BAM is defined as:
    
     \begin{equation} \label{eq:2}
		  x_1' = x_1 + B', \quad x_1' \in \mathbb{R}^{H \times W \times 32}.
    \end{equation}

	\subsection{MobileViT Block}
	\label{ssec:MV}
	MobileViT is a lightweight transformer that was proposed in \cite{mehta2021mobilevit}. The main component of MobileViT is the MobileViT Block (MVB), which can effectively encode global information while also implicitly incorporating convolution-like properties, as shown in the left gray box in Fig. \ref{fig:mymodel}. 
 
    Regarding local representation, the MVB contains a $3\times3$ convolutional layer that encodes local spatial information, as well as a point-wise convolution that projects the tensor to a high-dimensional space. In terms of global representation, the MVB first unfolds the tensor into non-overlapping flattened patches with each patch having a height and width of 4. Multiple transformers ($N$ transformers) are then applied to learn global representations, with $N$ being set to 2, 4, and 3 for $\rm{E_1}$, $\rm{E_2}$ and $\rm{E_3}$, respectively. These patches are then folded back into a feature map. Finally, the fusion block is a residual structure that can fuse the input and global representation features and restore the feature map to the input dimension using a $3\times3$ convolution.

	\subsection{Loss function}
	\label{ssec:Loss}
		
	Due to a small percentage of the pixels belonging to the cracks in crack images, serious data imbalance problem exists in crack segmentation. As a result, the model will primarily learn the features of non-crack samples with a large number of pixels.
 
    The weighted binary cross entropy loss function $L_{wbce}$ and dice loss function $L_{dice}$ \cite{milletari2016v} are widely used to address above problem. The weighted binary cross entropy loss can be defined as
	\begin{equation} \label{eq:3}
        \begin{split}
        L_{wbce} = &-\omega_0 y_i\sum\nolimits_i^N \log \hat{y}_i \\ 
                &-\omega_1 (1-y_i) \sum \nolimits_i^N \log (1-\hat{y}_i),
        \end{split}
	\end{equation}
    and the dice loss can be defined as
	\begin{equation} \label{eq:4}
		L_{dice} = 1 - \frac{2 \sum\nolimits_i^N y_i \hat{y}_i + 1}{\sum\nolimits_i^N y_i + \sum\nolimits_i^N \hat{y}_i + 1},
	\end{equation}
    where $y_i \in \{0, 1\}$ denotes the ground truth and $\hat{y}_i \in [0, 1]$ denotes the prediction of all pixels. Furthermore, $\omega_0$ and $\omega_1$ denote the loss weights for crack and non-crack pixels, respectively. We set $\omega_0 = \frac{\left| Y_- \right|}{\left| Y \right|}$, $\omega_1 = \frac{\left| Y_+ \right|}{\left| Y \right|}$, where $\left| Y \right|$,  $\left| Y_+ \right|$ and $\left| Y_- \right|$ denotes the total number of all pixels, all crack pixels and all non-crack pixels for an input image. Finally, the loss function can be defined as:
    

	\begin{equation} \label{eq:5}
		Loss = L_{wbce} + L_{dice}.
	\end{equation}

	\section{Experiments and results}
	\label{sec:experiments}
	
	\subsection{Datasets and Metrics}
	\label{ssec:Datasets}
	
	We evaluated the proposed network with two publicly available datasets: Crack500 \cite{yang2019feature,zhang2016road} and DeepCrack \cite{liu2019deepcrack}. The Crack500 dataset contains 3368 pavement crack images of 640 × 360 pixels. This dataset is divided into a training set of 1896 images, a validation set of 348 images, and a test set of 1124 images. And DeepCrack dataset contains 537 images with a size of 544 × 384 pixels, which is divided into 300 images as the training data, and 237 images as the test data.
	
	\begin{figure*}[htbp]
		\centerline{\includegraphics[scale=0.2]{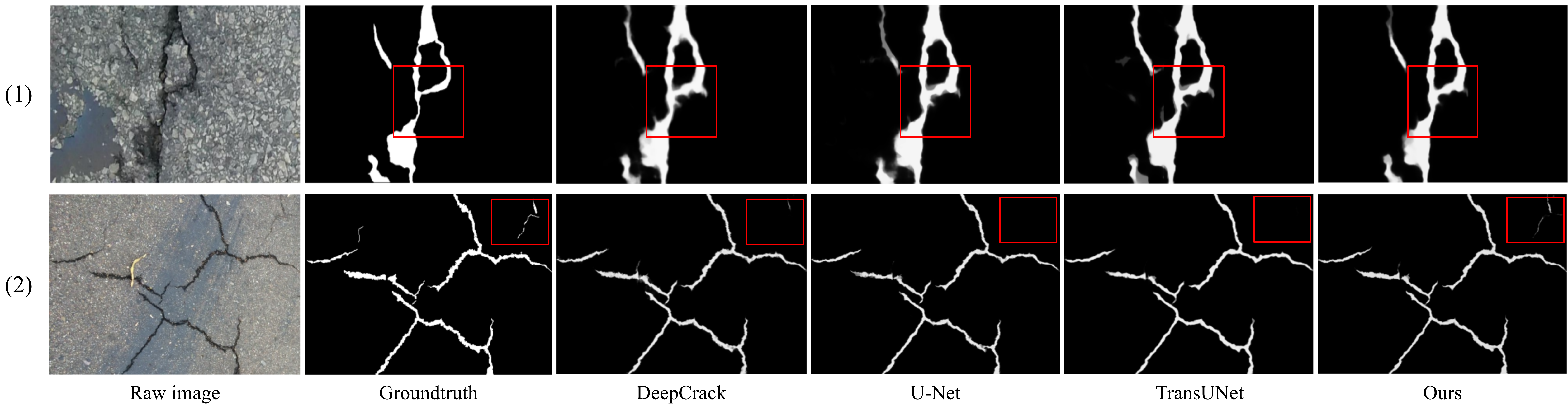}}
		\caption{Visual comparison of the segmentation results selected from (1)Crack500 and (2)DeepCrack. And our method can get clearer boundary in the first row, and capture more details in the second row than other methods.}
		\label{fig:Samples}
	\end{figure*}
  
        \begin{table*}
        \centering
        \caption{Evaluation metrics of competing methods on Crack500 and DeepCrack.}
        \label{tab:001}
        \belowrulesep=0pt
        \aboverulesep=0pt
        \scalebox{1.1}{
        \begin{tabular}{c|ccccccc}
        \Xhline{0.4pt}
        \multirow{2}{*}{$\textbf{Methods}$} & \multicolumn{3}{c}{Crack500} & \multicolumn{3}{c}{DeepCrack} & \multirow{2}{*}{$\bm{Params}$}\\
        \cmidrule(r){2-4}\cmidrule(r){5-7}
         & $\bm{Pr}$ & $\bm{Re}$ & $\bm{F1}$ & $\bm{Pr}$ & $\bm{Re}$ & $\bm{F1}$ &  \\
        \Xhline{0.4pt}
        
        U-Net \cite{ronneberger2015u} & 63.9\% & 68.4\% & 66.1\% & 85.0\% & 81.2\% & 83.0\% & 13.34M\\
        Deepcrack \cite{liu2019deepcrack} & 62.8\% & 67.6\% & 65.1\% & 86.1\% & $\textbf{86.9\%}$ & 86.5\% & 14.72M\\
        TransUNet \cite{chen2021transunet} & 60.5\% & 69.6\% & 64.7\% & 84.1\% & 83.4\% & 83.7\% & 101.19M\\
        Ours & $\textbf{69.1\%}$ & $\textbf{78.0\%}$ & $\textbf{73.3\%}$ & $\textbf{89.8\%}$ & $\textbf{86.9\%}$ & $\textbf{88.3\%}$ & 22.88M\\
 
        \end{tabular}
        }
        \end{table*}

	To fairly evaluate the performance of models, we utilize three common metrics: \emph{Precision} ($Pr$), \emph{Recall} ($Re$) and \emph{F1-score} ($F1$). They are calculated using True Positive, False Positives and False Negatives. And the parameter count ($Params$) is used to quantify the model’s complexity. 
	
	\subsection{Implementation Details}
	\label{ssec:Implementation}
	
	Our network is implemented on PyTorch \cite{paszke2017automatic} and the results are carried out on a single GPU system with Nvidia TITAN Xp 12 GB. Our network is based on U-Net, and after each convolution operation, batch normalization and ReLu are used. In training, we set batch size as 2 and an initial learning rate as $1\mathrm{e}{-4}$ for 100 epochs. We employ the Adam as the optimizer algorithm to update the network parameters.
	
	During the training phase, we resize the original image to 256 × 256 pixels, and the data augmentation methods \cite{info11020125} include flipping, rotation, and color jitter. In the inference phase, we crop the image into patches of 256 × 256 pixels. Finally, the predicted results of each patch are stitched together to generate a complete prediction image. 
	
        \subsection{Comparison with other methods}
	\label{ssec:Comparision}
	
	The experimental comparison results are shown in Table \ref{tab:001}. As can be seen, our method achieves the highest precision, recall, and F1-score. Specifically, on Crack500, the F1-score of U-Net, Deepcrack, and TransUNet falls short of our method by 7.2\%, 8.2\%, and 8.6\%, respectively. On DeepCrack, our method outperforms U-Net, Deepcrack, and TransUNet by 5.3\%, 1.8\%, and 4.6\%, respectively. It is worth noting that our model, unlike TransUNet, despite being equipped with ViT, still has appropriate parameters to strike a balance between model size and performance. To visually represent our results, a sample from each dataset is shown in Fig. \ref{fig:Samples}.

        \subsection{Ablation studies}
	\label{ssec:ablation}

    The ablation studies results are shown in Table \ref{tab:002}, where the baseline is U-Net \cite{ronneberger2015u}. As shown in Table \ref{tab:002}, DRB improves both precision and recall over baseline because it can capture more detailed local information than the conventional convolution layer. And v2, which can encoder global information by MVB, improves 1.7\% in precision and 1.5\% in recall than v1. Due to MVB being a lightweight transformer, v2 only adds 5.19M parameters than v1. In addition, BAM improves recall by 2.5\% over v2 with the almost same precision while only 0.01M parameters as the cost, indicating that crack boundary features can help determine the shape and location of cracks.

        \begin{table}[!htbp]
		\caption{Ablation studies results on DeepCrack.}\label{tab:002} \centering
		\begin{tabular}{c|c|c|c|c}
			\hline
			\textbf{Methods} & $\bm{Pr}$ & $\bm{Re}$ & $\bm{F1}$ & $\bm{Params}$\\
			\hline
                Baseline(B) & 85.0\% & 81.2\% & 83.0\% & 13.34M\\
			B+DRB(v1) & 87.7\% & 82.9\% & 85.2\% & 17.68M\\
			v1+MVB(v2) & 89.4\% & 84.4\% & 86.8\% & 22.87M\\
                v2+BAM(v3) & $\textbf{89.8\%}$ & $\textbf{86.9\%}$ & $\textbf{88.3\%}$ & 22.88M\\
		\end{tabular}
	\end{table}

    \vspace{-5pt}
	\section{Conclusion}
	\label{sec:conclusion}
	
	This paper proposes a novel convolutional-transformer network for crack segmentation, which includes dilated residual blocks, a boundary awareness module, and MobileViT blocks. The first two modules are convolutional blocks specially designed for slender crack segmentation, and the last module associates the global information of the image with only a few additional parameters. Extensive experiments have shown the superiority of our method.
	
	We hope that this study will inspire further research on convolutional-transformer networks for crack segmentation. In the future, we are interested in improving our model to get more valuable information from crack images, such as crack extension path and depth, in order to better monitor structural health.
	
	
	\vfill\pagebreak
	
	\bibliographystyle{IEEEbib}
	\bibliography{strings,refs}
	
\end{document}